\documentclass[letterpaper, 10 pt, conference]{ieeeconf}  

\IEEEoverridecommandlockouts                              

\usepackage{graphics} 
\usepackage{epsfig} 
\usepackage{mathptmx} 
\usepackage{times} 
\usepackage{amsmath} 
\usepackage{amssymb}  
\usepackage{subcaption}
\usepackage{multirow}
\usepackage{url}
\title{\LARGE \bf
Bi-LAT: Bilateral Control-Based Imitation Learning via Natural Language and Action Chunking with Transformers
}

\author{Takumi Kobayashi$^{1\dag}$, Masato Kobayashi$^{1,2,3\dag*}$, Thanpimon Buamanee$^{1}$, Yuki Uranishi$^{1,2}$ 
\thanks{
${\dag}$ Equal Contribution,
$^{1}$ Graduate School of Information Science and Technology, The University of Osaka, $^{2}$ D3 Center, The University of Osaka, $^{3}$ Graduate School of Maritime Sciences, Kobe University, * corresponding author: kobayashi.masato.cmc@osaka-u.ac.jp}
}

\begin{document}

\maketitle
\thispagestyle{empty}
\pagestyle{empty}

\begin{abstract}
We present Bi-LAT, a novel imitation learning framework that unifies bilateral control with natural language processing to achieve precise force modulation in robotic manipulation.
Bi-LAT leverages joint position, velocity, and torque data from leader-follower teleoperation while also integrating visual and linguistic cues to dynamically adjust applied force. 
By encoding human instructions such as ``softly grasp the cup'' or ``strongly twist the sponge'' through a multimodal Transformer-based model, Bi-LAT learns to distinguish nuanced force requirements in real-world tasks.
We demonstrate Bi-LAT's performance in (1) unimanual cup-stacking scenario where the robot accurately modulates grasp force based on language commands, and (2) bimanual sponge-twisting task that requires coordinated force control.
Experimental results show that Bi-LAT effectively reproduces the instructed force levels, particularly when incorporating SigLIP among tested language encoders. 
Our findings demonstrate the potential of integrating natural language cues into imitation learning, paving the way for more intuitive and adaptive human–robot interaction.
For additional material, please visit the website: \url{https://mertcookimg.github.io/bi-lat/}

\end{abstract}

\section{INTRODUCTION}
In today’s rapidly evolving landscape of robotics, integrating advanced manipulation capabilities with social intelligence is pivotal to shaping our hybrid future. Robotic manipulation plays a central role in a range of human-centric applications—from cooking assistance \cite{liu2022robot} and eldercare \cite{liu2024state} to dynamic interactive scenarios \cite{tsai2022sanitizerbot}—where robots are envisioned as collaborative partners. While conventional industrial robots excel at executing pre-programmed, repetitive actions, the emerging generation of service and collaborative robots must adapt to dynamic environments and handle objects with diverse shapes and material properties. This transformation calls for learning-based approaches that replicate complex, human-like manipulation strategies, including precise force regulation, and that are seamlessly integrated into broader human-robot interaction frameworks.

Among these methods, imitation learning (IL) has emerged as an effective method for enabling robots to acquire new skills directly from imitating human demonstrations \cite{zare2024survey}. By circumventing the need to explicitly design a reward function, IL addresses tasks that are difficult to formalize, such as delicate assembly \cite{ankile2024juicer} or surgical procedures \cite{arduini2024learning}, and distills expert knowledge into a parameterized policy. However, most existing IL frameworks in robotics traditionally assume \textit{unilateral} control-based teleoperation setups, in which human demonstrators only provide position commands. Although these systems are relatively straightforward to implement, they often struggle to capture the rich force interactions essential for manipulating fragile or deformable objects.
\begin{figure}[t]
\centering
\includegraphics[keepaspectratio, width=1\linewidth]{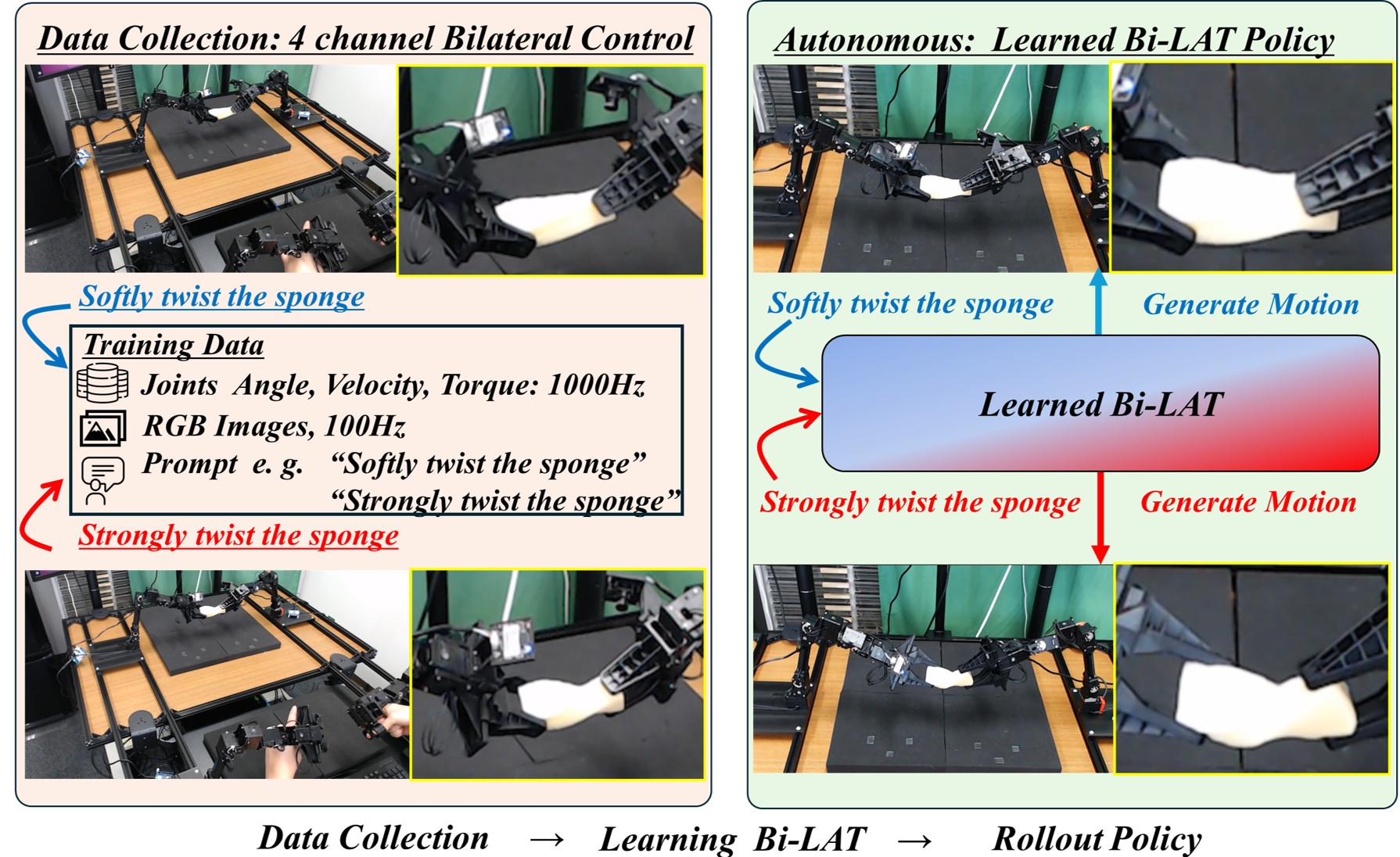}
\caption{Bilateral Control-Based Imitation Learning via Natural Language and Action Chunking with Transformers}
\label{fig:teser}
\end{figure}

In parallel with advances in imitation learning, the emergence of foundation models and large language models (LLM) has dramatically broadened the horizons of robotics research \cite{brohan2023can, kannan2024smart}. Although these models have achieved notable success in planning and semantic perception, they often fall short when it comes to the fine-grained control of force and torque, which are critical factors for safe and effective object manipulation.

To bridge this gap in force-centric manipulation, many researchers have turned to bilateral control, wherein both position and force information are exchanged between the human demonstrator and the robot \cite{kobayashi2025alpha}. Bilateral control-based teleoperation enables the demonstrator to feel contact forces in real time, capturing more accurate physical interactions for training data. Indeed, prior studies have demonstrated that bilateral control-based imitation learning enables better generalization across objects with different hardness and weights compared to unilateral setups \cite{adachi2018imitation, hayashi2022independently, sakaino2022imitation}. Recently, bilateral control-based imitation learning via action chunking with transformers (Bi-ACT)~\cite{buamanee2024bi} showed that combining bilateral control information and visual inputs with Transformer-based models improves action accuracy. However, it does not yet incorporate natural language instructions specifying how much force to apply.

In this paper, we propose Bi-LAT: Bilateral Control-Based Imitation Learning via Natural Language and Action Chunking with Transformers. By leveraging multimodal Transformers that fuse robot information (joint angles, velocities, torques), visual information, and linguistic cues, Bi-LAT links everyday language instructions (e.g., ``softly grasp the cup'') directly to explicit force modulation during task execution. This integration enables robots to autonomously adjust force based on higher-level, human-readable commands rather than relying on numerically defined torque thresholds. We validate Bi-LAT on two real-world tasks, the unimanual cup-stacking task and the bimanual sponge-twisting task, both highlighting the necessity of precise force control.

Our main contributions are as follows:
\begin{itemize} \item We propose Bi-LAT, a bilateral control-based imitation learning framework that integrates robot position and force data, visual observations, and natural language instructions. 
\item We evaluate several language/text encoders in unimanual cup-stacking scenario and demonstrate that the SigLIP text encoder particularly excels at modulating grasp force according to language instructions.
\item We further demonstrate the effectiveness of Bi-LAT in bimanual setting, where coordinated force control is required to twist a deformable sponge, validating our method for more complex tasks. \end{itemize}

This paper consists of six sections. Section II explains  Related Works.
Section III proposes Bi-LAT method.
Sections IV and V confirm the method's efficiency based on the unimanual and bimanual experimental results.
Section VI concludes this paper.

\section{RELATED WORKS}
\subsection{Bilateral Control-Based Imitation Learning}
Bilateral control is used for data collection, which involves the remote operation of a follower robot in the environment, guided by a leader robot controlled by human~\cite{kobayashi2025alpha}.
This approach provides operators with force feedback during data collection, which is crucial for capturing motion data that closely aligns with human intuition.
These are achieved through position tracking and the use of action-reaction principles.

Many studies on bilateral control-based imitation learning employ neural network architecture such as Long Short-Term Memory (LSTM). For example, previous studies have demonstrated tasks like drawing straight lines along a ruler \cite{adachi2018imitation}, wiping a whiteboard with an eraser \cite{hayashi2022independently}, and thinly slicing cucumbers \cite{sakaino2022imitation}, all of which require precise force control for successful execution.
More recently, Transformers have gained traction in robotic control. Notably, Bi-ACT \cite{buamanee2024bi} integrates Action Chunking with Transformers (ACT) \cite{zhao2023learning} into bilateral control-based imitation learning. Bi-ACT predicts not only the next action but also sequences of actions, thereby addressing the issue of cumulative errors, a common challenge in imitation learning. Moreover, Bi-ACT extends bilateral control-based imitation learning by incorporating visual data, which enables flexible adaptation to objects with varying weights and shapes. Recent developments related to Bi-ACT include ALPHA-$\alpha$ \cite{kobayashi2025alpha}, a low-cost bimanual robotic platform, and the data augmentation method for bilateral control-based imitation learning with images (DABI) \cite{kobayashi2024dabievaluationdataaugmentation}, which enhances training data by downsampling both image and robot data to improve learning efficiency.

In this paper, we extend Bi-ACT by integrating natural language processing with a language/text encoder, creating a bilateral control-based imitation learning framework called Bi-LAT that generates actions based on human language instructions while effectively leveraging force information.

\subsection{Imitation Learning Using Language Models} 
Traditionally, robots have been controlled through explicit programming, with commands defined as specific control values such as current, torque, and position. However, the advent of language models has revolutionized this paradigm, enabling human operators to provide instructions using natural language \cite{stepputtis2020language, zhou2024language}.
Large language models further enhance a robot’s ability to interpret natural language instructions and generate structured action plans, such as determining the correct sequence of actions to execute a task \cite{brohan2023can, kannan2024smart}. The emergence of vision-language models (VLM) has facilitated the integration of visual and textual information, allowing robots to process and interpret images and videos alongside language commands \cite{gao2024physically, li2024learning}. 
Collectively, these advancements represent a paradigm shift from rigid, pre-programmed instructions to flexible, context-aware, and human-intuitive interaction mechanisms.

In this paper, we proposed Bi-LAT. By linking linguistic information with the haptic feedback provided by bilateral control, Bi-LAT enables robots to execute tasks with precise force modulation instructed by human operators.

\begin{figure}[t]
\centering
\includegraphics[keepaspectratio, width=1\linewidth]{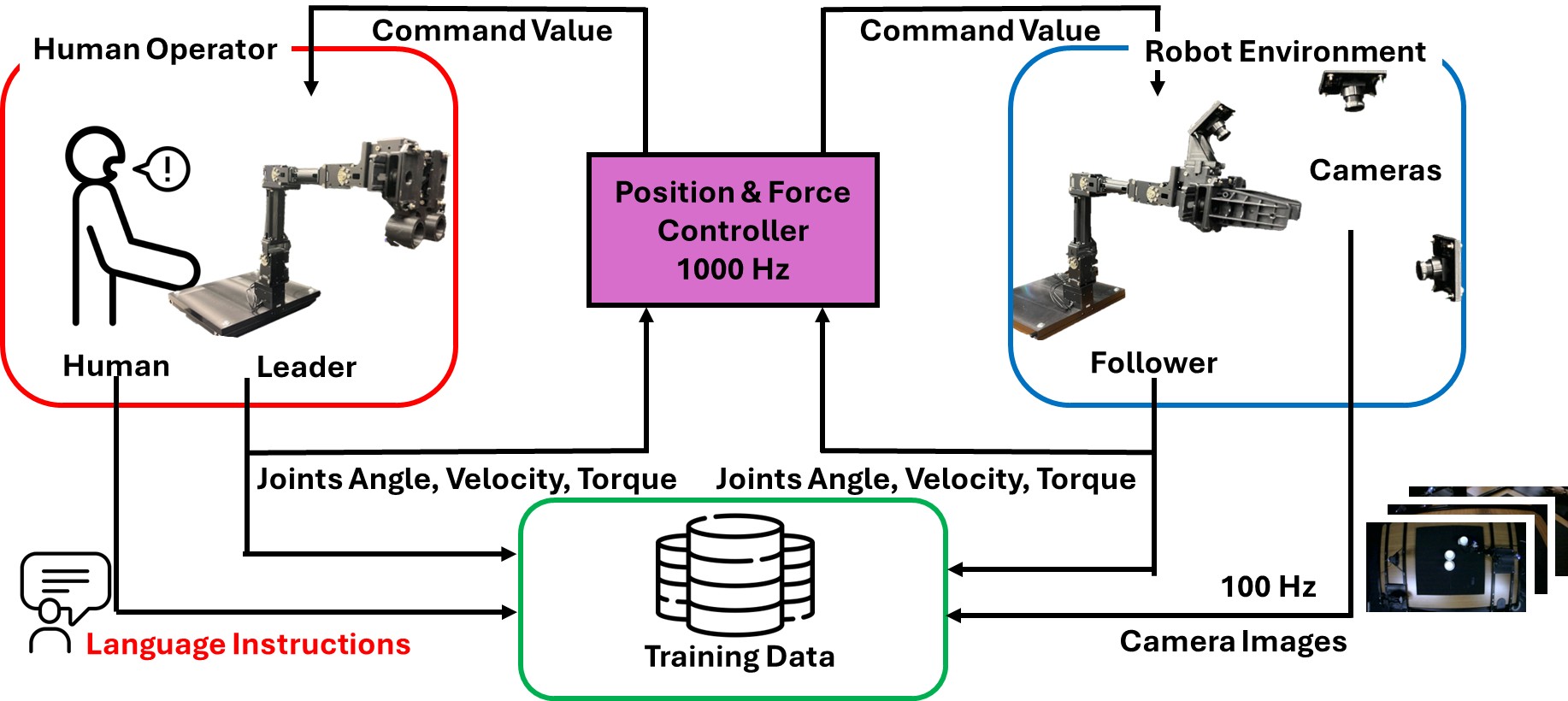}
\caption{Bi-LAT: Data Collection}
\label{fig:bi-lat_datacollection}
\end{figure}

\begin{figure*}[t]
\centering
\includegraphics[keepaspectratio, width=1\linewidth]{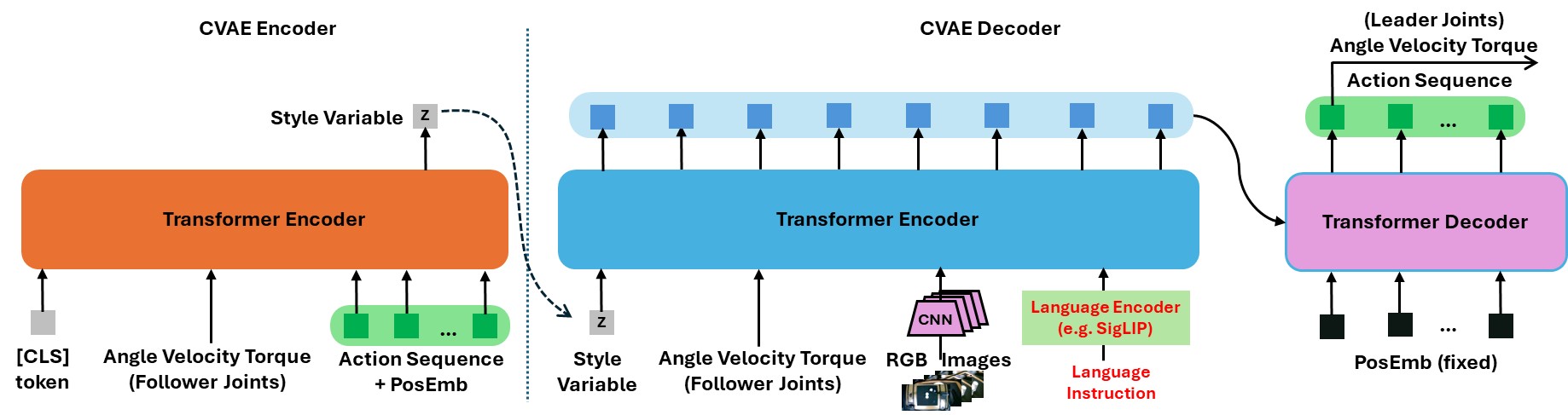}
\caption{Bi-LAT: Learning Model}
\label{fig:bi-lat-model}
\end{figure*}
\section{Bi-LAT: Bilateral Control-Based Imitation Learning via Natural Language and Action Chunking with Transformers}
\begin{figure}[t]
\centering
\includegraphics[keepaspectratio, width=0.85\linewidth]{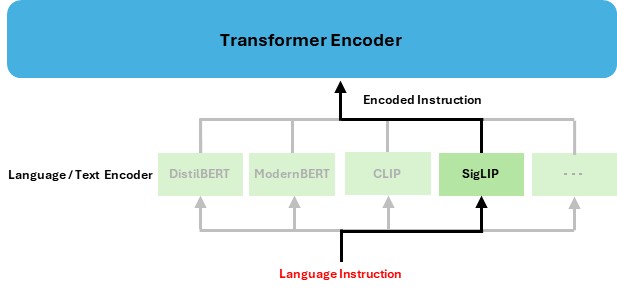}
\caption{Language Encoder}
\label{fig:Encoding}
\end{figure}

\subsection{Overview}\label{sec:bi-lat_overview}
Bi-ACT leverages bilateral control to perform robot control using both positional and force information, enabling flexible adaptation even when handling previously unseen objects.
Additionally, by incorporating image data from RGB cameras, the system achieves visual-based robot control.

Building on this foundation, the proposed method Bi-LAT integrates a language encoder into the framework to ensure that generated actions correspond to the force level specified by the operator, as shown in Fig.~\ref{fig:teser}.
Incorporating natural language processing techniques allows robots to execute tasks based on user-provided natural language instructions.

\subsection{Data Collection}\label{sec:bi-lat_datacollection}
Bi-LAT employs four-channel bilateral control method for data collection.
The key concept underlying bilateral control is that both the operator and the controlled target continuously share position, force, and other relevant information to achieve coordinated behavior.
This coordination is mathematically characterized by the equations:
\begin{equation}
\theta_l - \theta_f = 0
\label{eq:position} 
\end{equation}
\begin{equation}
\tau_l + \tau_f = 0
\label{eq:force}
\end{equation}
In these expressions, $\theta$ denotes joint angles and $\tau$ denotes torques, with the subscripts $\bigcirc_l$ and $\bigcirc_f$ indicating the leader and follower systems.
Encoders are used to capture joint angles.
Meanwhile, the torque response is estimated via disturbance observer (DOB) \cite{ohnishi1996motion} and reaction force observer (RFOB) \cite{murakami2002torque}.
This method facilitates the estimation of environmental reaction torques without force/torque sensors.

Bi-LAT extends the data collection process by incorporating natural language instructions provided by the operator, as shown in Fig.~\ref{fig:bi-lat_datacollection}. 
The operator teleoperates the follower robot via the leader robot, while natural language instructions describing the movements are recorded alongside robot joint data and camera images. This additional modality enables the system to later interpret these natural language cues for precise force control, effectively bridging the gap between human tactile perception and robotic actuation.
To further enhance the performance of the language model, we employ prompt engineering techniques that have proven effective with Contrastive Language–Image Pre-training (CLIP) \cite{radford2021learning}.
Previous studies have indicated that CLIP’s accuracy can degrade when processing ambiguous sentences or isolated words; however, appending the phrase ``a photo of a'' to the beginning of a prompt has been shown to improve performance. 
We also apply similar prompt engineering strategies during data collection.

\subsection{Language Encoder}\label{sec:LanguageModels}
As shown in Fig.~\ref{fig:Encoding}, we adopt several language encoders from CLIP \cite{radford2021learning} and Bidirectional Encoder Representations from Transformers (BERT) \cite{devlin2019bert} architectures to convert natural language instructions into distributed representations for processing by the Bi-LAT model. Specifically, we employ CLIP and its derivative, SigLIP \cite{zhai2023sigmoid}, from the CLIP series, along with DistilBERT \cite{sanh2019distilbert} and ModernBERT \cite{warner2024smarter} from the BERT series. CLIP-based models excel at integrating visual and textual cues, whereas BERT-based models are highly effective in natural language understanding. By leveraging the strengths of these architectures, our approach enables robust interpretation of language commands for precise force control in robotic tasks.

\begin{figure}[t]
\centering
\includegraphics[keepaspectratio, width=1\linewidth]{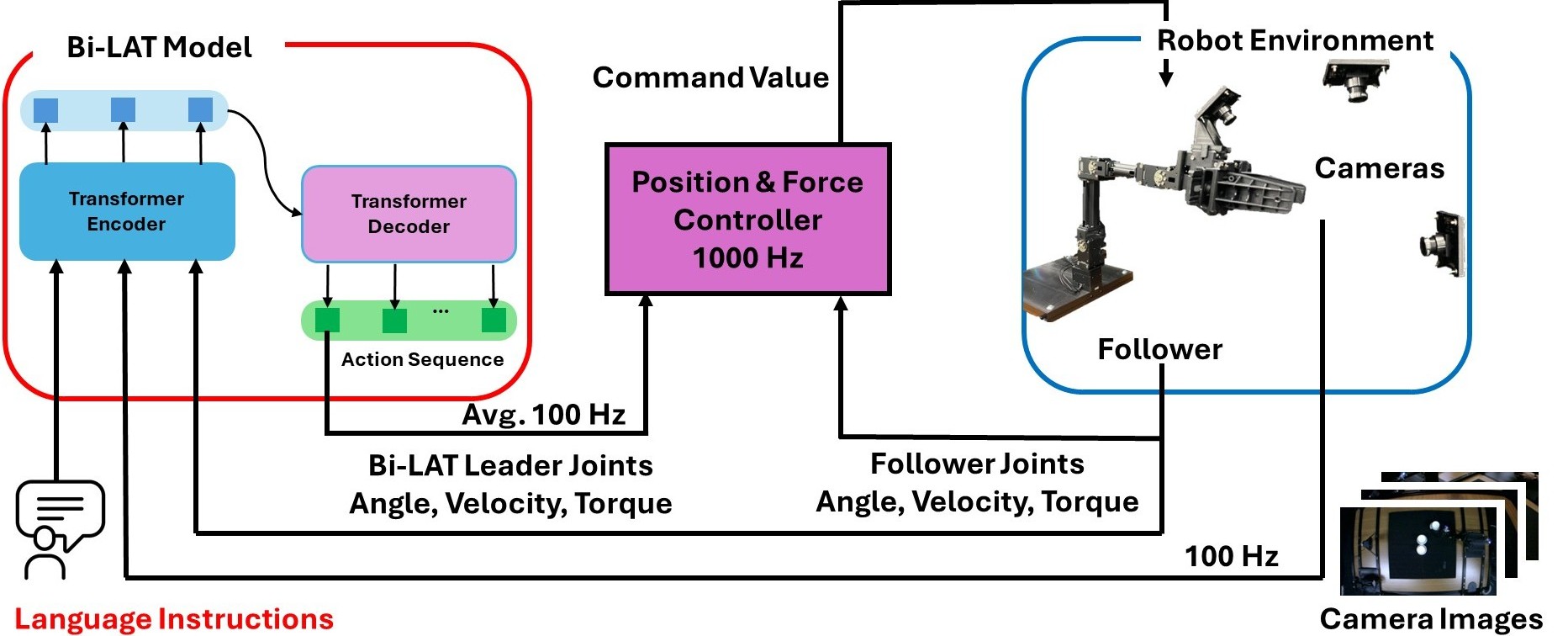}
\caption{Bi-LAT: Inference}
\label{fig:bi-lat_inference}
\end{figure}

\subsection{Learning Model}\label{sec:bi-lat_Model}
The Bi-LAT learning model is built upon a Transformer-driven Conditional Variational Autoencoder (CVAE) architecture, as illustrated in Fig.~\ref{fig:bi-lat-model}.
In this framework, the model receives the follower robot joints' angle, velocity, and torque data along with corresponding visual inputs and natural language instructions. Based on this multimodal input, the model outputs sequences of action chunks that specify the predicted joint angles, velocities, and torques of leader robots.
Natural language instructions are processed by a dedicated language encoder (LE), which converts the textual commands into fixed-length vector representations.
In this paper, we select four language encoders among many available options, namely DistilBERT, ModernBERT, CLIP, and SigLIP, as shown in Fig.~\ref{fig:Encoding}.
These linguistic features are integrated with visual features extracted from ResNet-18 and the follower robot's joint angles, velocities, and torques to form a comprehensive latent representation. This representation is then passed through the CVAE-based Transformer decoder, which predicts the future leader robot's actions as a series of action chunks.
The learning process is driven by minimizing the error between the predicted leader joint data and the corresponding ground-truth data collected during bilateral control teleoperation. 

\begin{figure}[t] 
    \centering \includegraphics[keepaspectratio, width=0.8\linewidth]{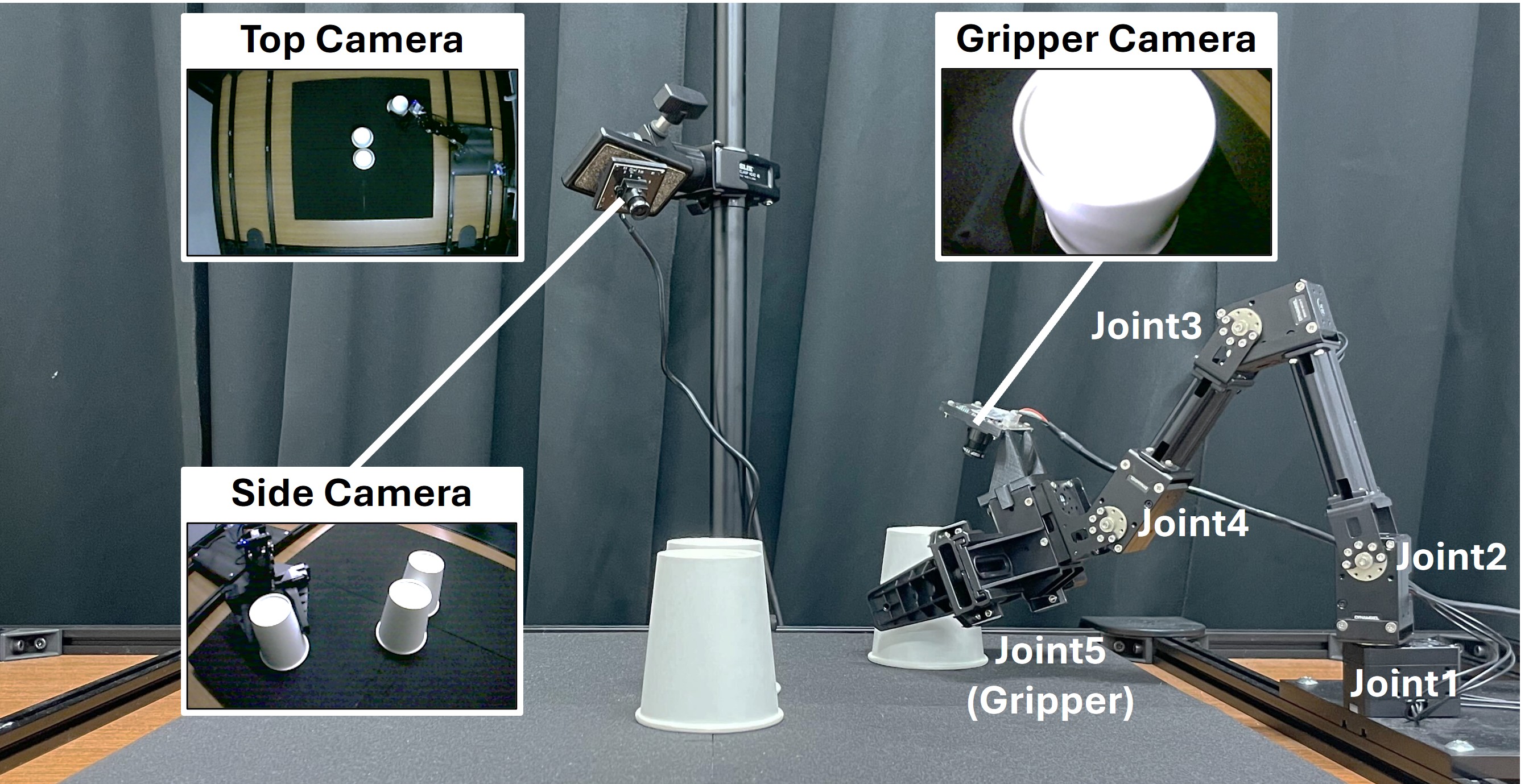} 
    \caption{Unimanual Experimental Environment} 
    \label{fig:ex1_hardware} 
\end{figure}

\subsection{Inference}\label{sec:bi-lat_inference}
Fig.~\ref{fig:bi-lat_inference} illustrates the operational diagram for Bi-LAT inference.
During inference, the Bi-LAT model receives the most recent joint data from the follower robot along with corresponding camera images and natural language instruction specifying the desired action.
Based on this input, the model predicts the subsequent action chunk of leader robot's joints information.
Bi-LAT outputs action data that include joint angles, velocities, and torques for each joint.
These outputs are then converted into the requisite current for each joint by calculations performed by the bilateral control system. 

\section{Unimanual Experiment}
\subsection{Hardware}
Fig.~\ref{fig:ex1_hardware} shows the environment in which the unimanual experiment was conducted.
We used the OpenManipulator-X from ROBOTIS.
This manipulator features four joints (Joint1 to Joint4) for arm rotation, while Joint5 is dedicated to gripper actuation. 
In the unimanual experiment, two robot arms were used: one configured as the leader robot and the other as the follower robot.
For collecting image data, we used the ELP USB camera (model ELP-USBFHD08S-L36), which captures RGB images at a resolution of 640×360 pixels.
Three of these cameras were deployed overhead, on the side, and on the gripper to ensure comprehensive visual coverage.

\begin{figure}[t]
    \centering
    \includegraphics[keepaspectratio, width=1\linewidth]{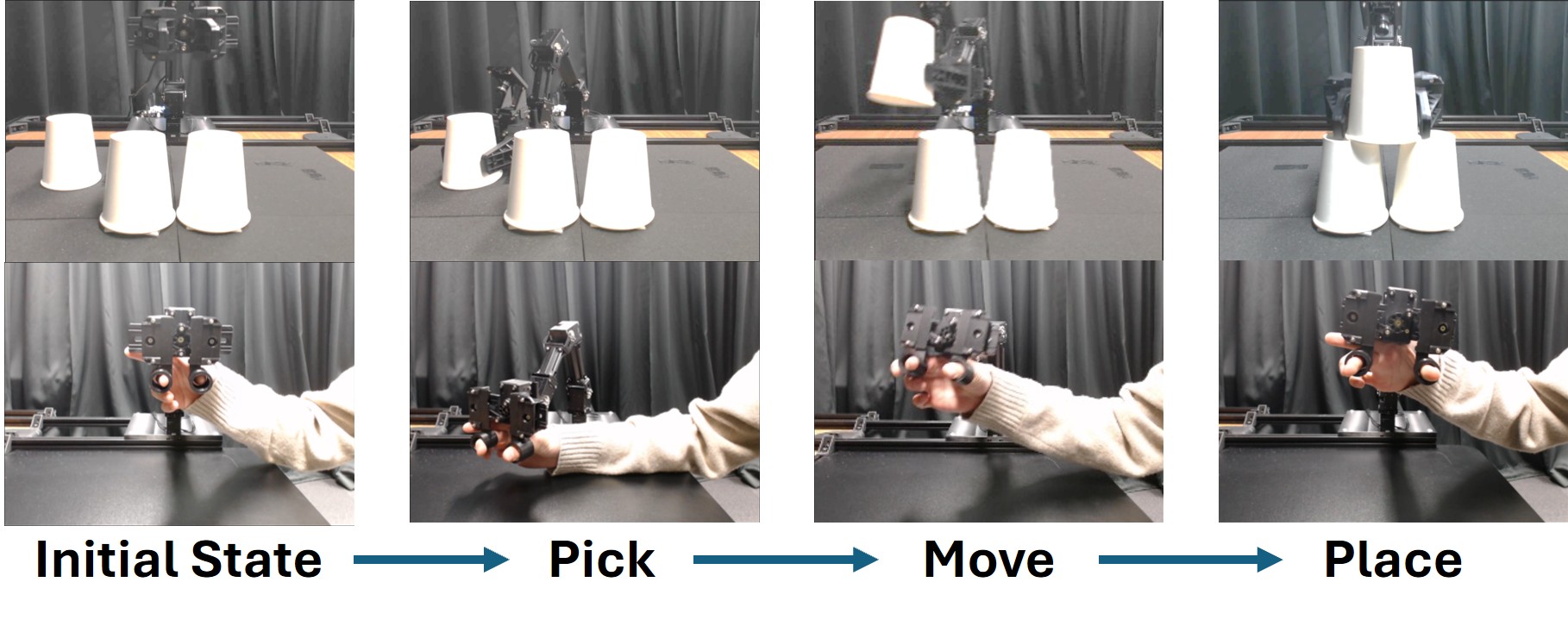}
    \caption{Detailed Cup-Stacking Task} 
    \label{fig:task_summary} 
\end{figure}

\subsection{Experimental Conditions}
To verify that language instructions are correctly associated with the robot’s joint position and torque values, we conducted a cup-stacking experiment using three paper cups.
The deformable nature of these cups allowed us to test whether the robot could autonomously control its grasping force based on natural language instructions.

As depicted in Fig.~\ref{fig:task_summary}, the cup-stacking task is divided into three stages:
\begin{itemize}
\item \textit{Pick}: The robot moves from its initial position to the front of the cup, opens its gripper, moves forward, and then closes its gripper to grasp the cup. 
\item \textit{Move}: While maintaining its grip on the cup, the robot moves to the apex of the two centrally placed cups. 
\item \textit{Place}: The robot opens its gripper in the center of the two base cups to release and place the cup.
\end{itemize}

To examine the effect of varying force requirements, we collected demonstrations for two distinct instructions: 
\begin{itemize}
    \item \textit{``Softly grasp the cup''} (lighter grip force)
    \item \textit{``Strongly grasp the cup''} (firmer grip force)
\end{itemize}

Because the paper cups were deformable, these instructions produced noticeably different torque profiles, especially during the grasping phase.

\begin{figure}[t] 
    \centering 
    \includegraphics[keepaspectratio, width=1\linewidth]{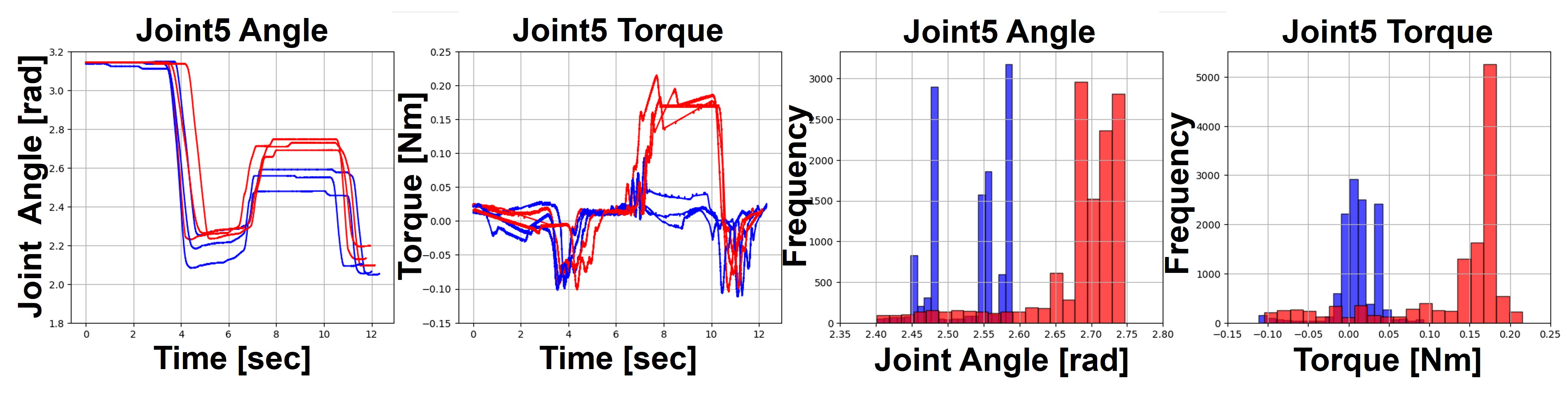} 
    \caption{Training Data of Joint5 (Gripper)} 
    \label{fig:trainingdata} 
\end{figure}
\subsection{Training Setup}
Joint angles, angular velocities, and torques were recorded from the leader and follower robots using the four-channel bilateral control system, operating at a control frequency of 1000 Hz.
This resulted in 15-dimensional joint data (3 values × 5 joints) per robot, and 30-dimensional joint data in total across both robots, capturing detailed motor-level dynamics essential for imitation learning.
Simultaneously, RGB images were captured from three cameras at a frequency of 100 Hz.

As shown in Fig.~\ref{fig:trainingdata}, each demonstration was paired with a natural language instruction.
Among the six demonstrations, three involved ``softly grasp the cup'', while the other three involved ``strongly grasp the cup.''
Histograms for Joint5 (Gripper) angle and torque during the grasping phase support these findings.
Under the ``softly grasp the cup'' instruction, most joint angles are distributed between 2.45 and 2.60 rad, with torques between 0 and 0.05 Nm. 
Under the ``strongly grasp the cup'' instruction, most joint angles concentrate between 2.65 and 2.75 rad, with torques between 0.15 and 0.20 Nm.

Moreover, to augment the training data, we employed the DABI~\cite{kobayashi2024dabievaluationdataaugmentation}.
DABI is particularly effective when high-frequency robot sensor data and lower-frequency image data are collected simultaneously.
By downsampling the 1000 Hz control data to 100 Hz, DABI expanded the original six demonstrations to 60, thereby increasing the dataset size by a factor of ten.
For additional details, please refer to \cite{kobayashi2024dabievaluationdataaugmentation}.
Based on this dataset, we trained Bi-ACT, and Bi-LAT using DistilBERT, ModernBERT, CLIP text encoder, and SigLIP text encoder.
We configured Bi-LAT models with 4 encoder layers and 7 decoder layers.

\subsection{Experiment Result}

\begin{table}[t]
    \centering
    \caption{Cup-Stacking Success Rates}
    \label{tab:cup_stack_result}
    \scalebox{0.9}{
    \begin{tabular}{lccc} \hline
        \textbf{Method} & \textbf{Instruction} & \textbf{Success Rate} & \textbf{Force Accuracy} \\ \hline\hline
        Bi-ACT(None) & No Instruction & 5/5(100\%) & $\times$ \\ \hline
        \multirow{2}{*}{Bi-LAT(DistilBERT)} & Softly  & 5/5(100\%) & \multirow{2}{*}{$\bigtriangleup$} \\
        & Strongly  & 5/5(100\%) & \\ \hline
        \multirow{2}{*}{Bi-LAT(ModernBERT)} & Softly  & 5/5(100\%) & \multirow{2}{*}{$\times$} \\
        & Strongly  & 5/5(100\%) & \\ \hline
        \multirow{2}{*}{Bi-LAT(CLIP)} & Softly  & 5/5(100\%) & \multirow{2}{*}{$\times$} \\
        & Strongly  & 5/5(100\%) & \\ \hline
        \multirow{2}{*}{Bi-LAT(SigLIP)} & Softly  & 5/5(100\%) & \multirow{2}{*}{$\bigcirc$} \\
        & Strongly  & 5/5(100\%) & \\
        \hline
    \end{tabular}
    }
\end{table}

\begin{figure}[t] 
    \centering 
    \includegraphics[keepaspectratio, width=1\linewidth]{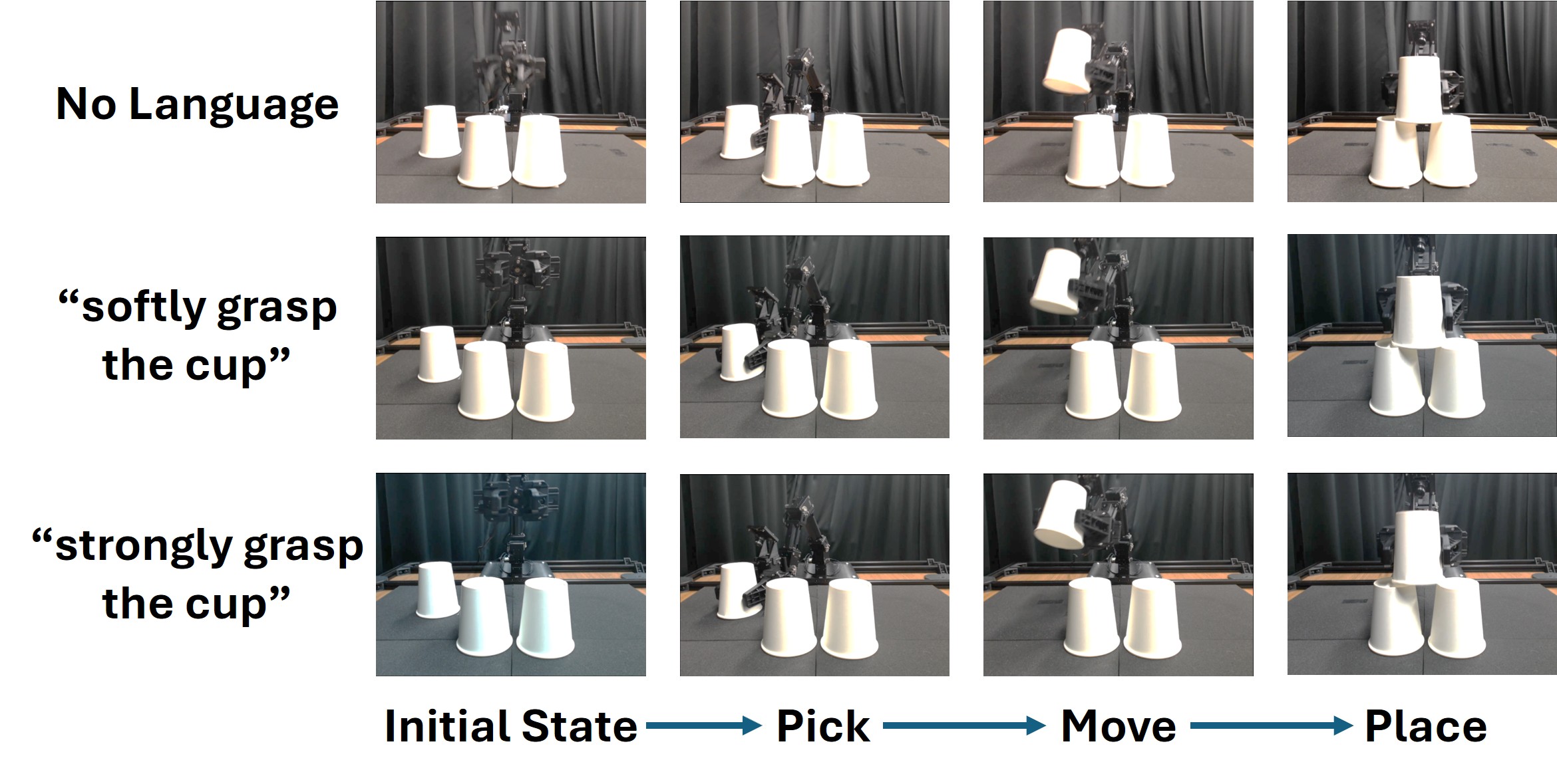} 
    \caption{Autonomous Execution using Bi-ACT (No Language Instructions) and using Bi-LAT (with SigLIP)}
    \label{fig:ex1_result}
\end{figure}

\begin{figure}[t]
    \begin{minipage}[b]{1\linewidth}
        \centering
        \includegraphics[keepaspectratio, width=1\linewidth]{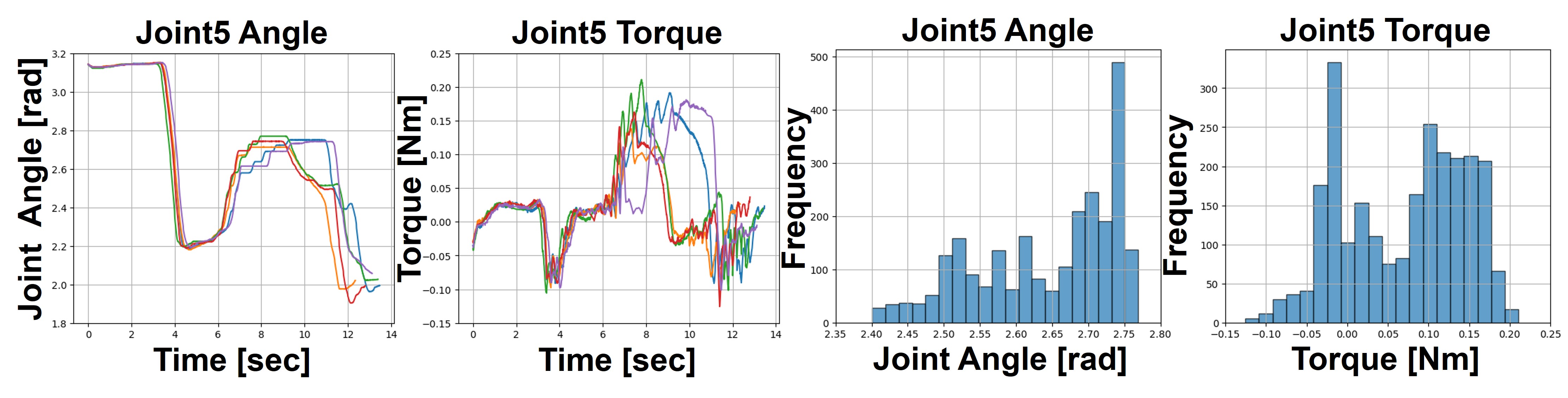}
        \subcaption{Bi-ACT (None)}
    \end{minipage}
    \begin{minipage}[b]{1\linewidth}
        \centering
        \includegraphics[keepaspectratio, width=1\linewidth]{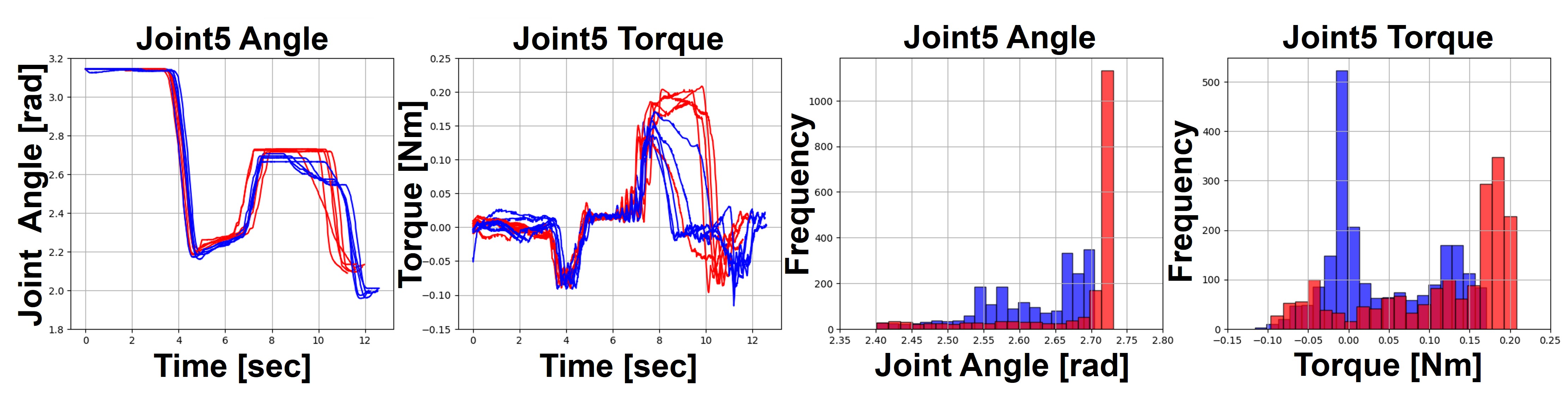}
        \subcaption{Bi-LAT (DistilBERT)}
    \end{minipage}
    \begin{minipage}[b]{1\linewidth}
        \centering
        \includegraphics[keepaspectratio, width=1\linewidth]{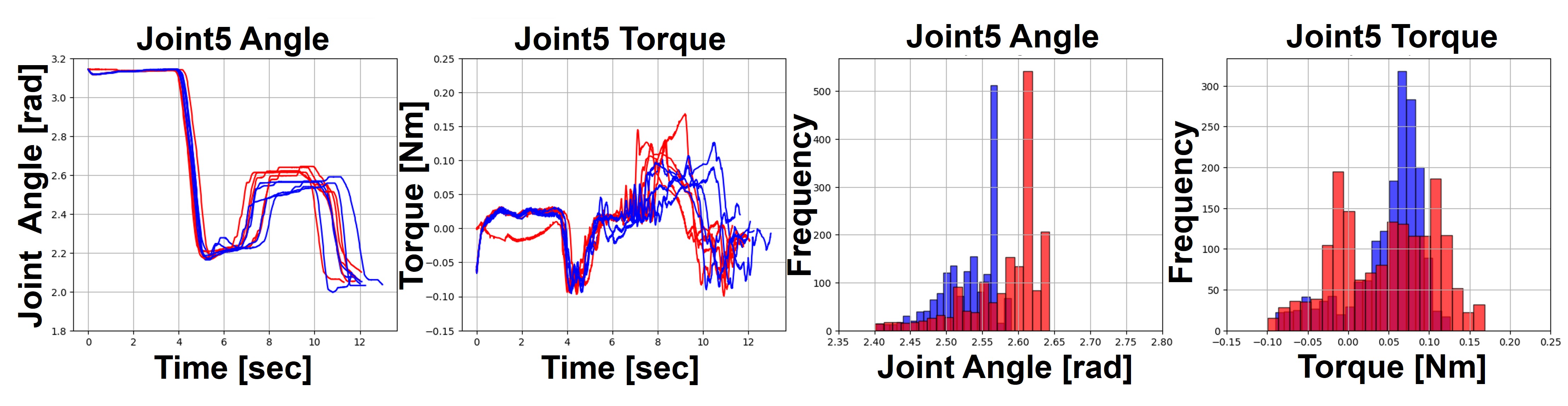}
        \subcaption{Bi-LAT (ModernBERT)}
    \end{minipage}
    \begin{minipage}[b]{1\linewidth}
        \centering
        \includegraphics[keepaspectratio, width=1\linewidth]{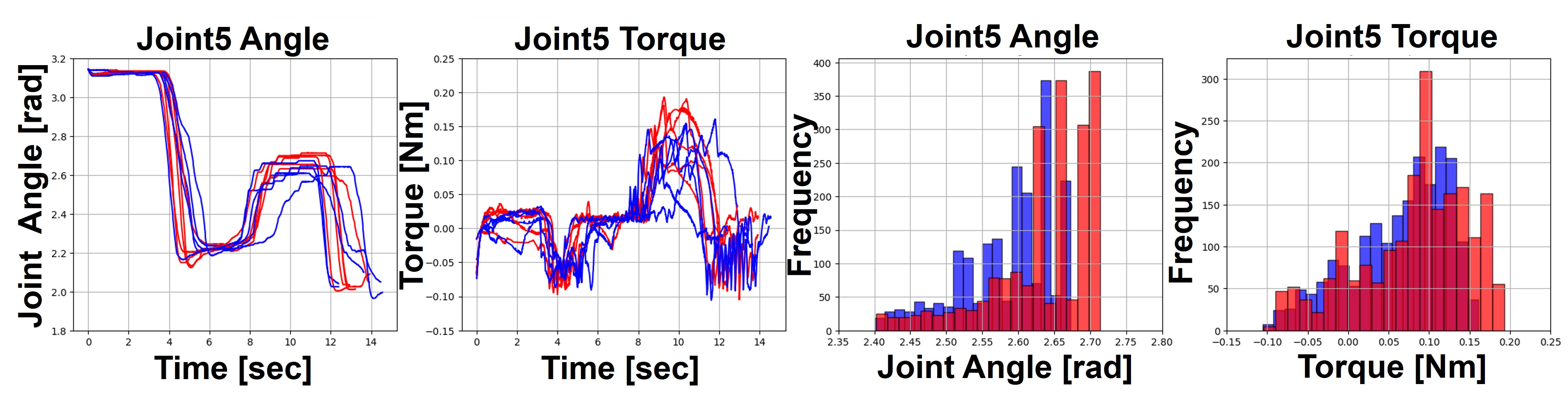}
        \subcaption{Bi-LAT (CLIP)}
    \end{minipage}
    \begin{minipage}[b]{1\linewidth}
        \centering
        \includegraphics[keepaspectratio, width=1\linewidth]{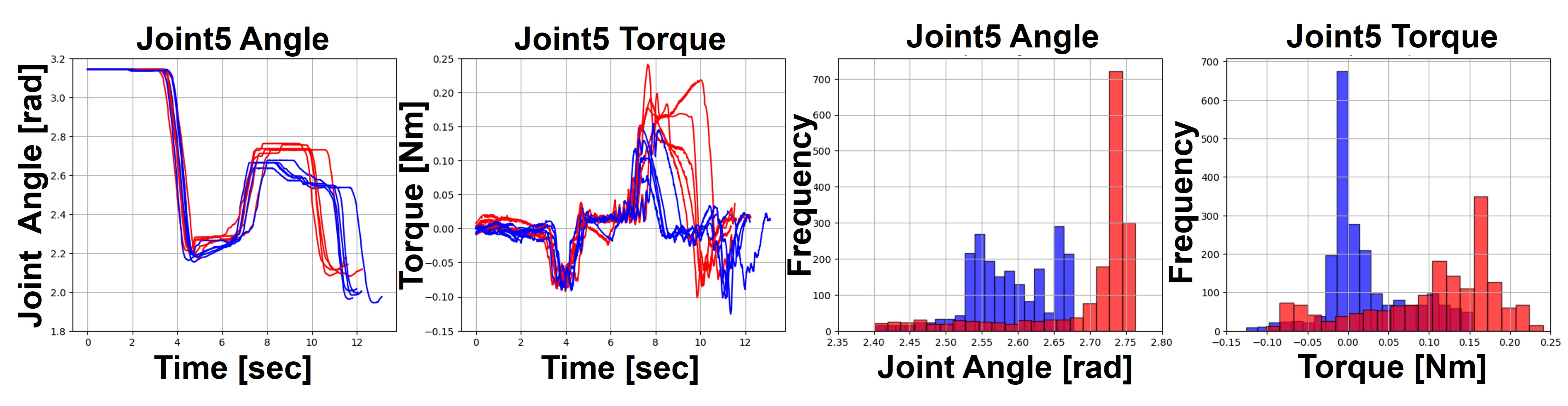}
        \subcaption{Bi-LAT (SigLIP)}
    \end{minipage}
    \caption{Results of Gripper Joint5 Angle and Torque}
    \label{fig:cup_exp_all}
\end{figure}
\begin{figure*}[t] 
    \centering 
    \includegraphics[keepaspectratio, width=1\linewidth]{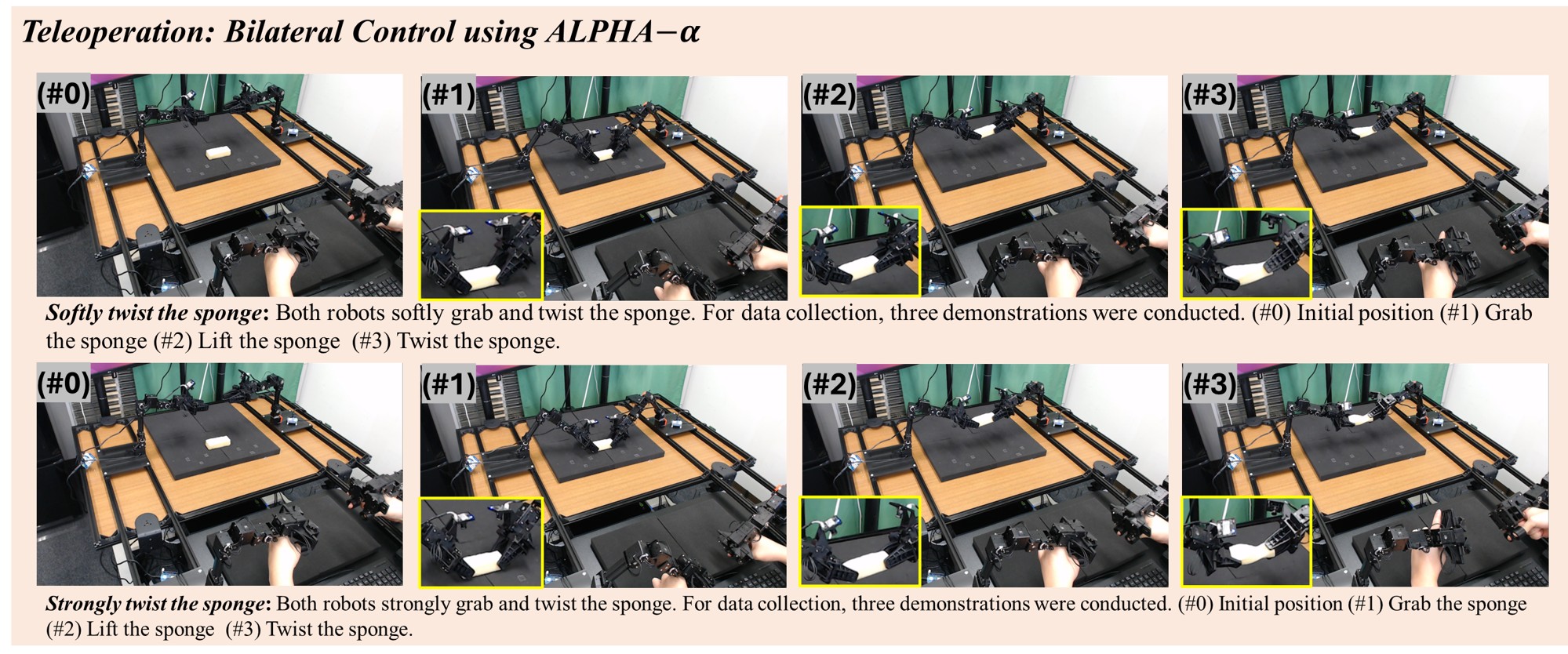} 
    \caption{Data Collection of Sponge-Twisting Task using ALPHA-$\alpha$} 
    \label{fig:ex2_datacollection} 
\end{figure*}
As shown in Table~\ref{tab:cup_stack_result}, the cup-stacking task was successfully achieved in 100\% of trials.
Fig.~\ref{fig:ex1_result} shows the results of autonomous execution using the conventional Bi-ACT (no language instructions) and Bi-LAT with SigLIP as the language encoder.

In the conventional Bi-ACT model, which was trained without language inputs, data from both ``softly'' and ``strongly'' demonstrations were merged into a single training set. As a result, the model treated both force modalities as equivalent aspects of the same task. As shown in Fig.~\ref{fig:cup_exp_all}(a), when no language instruction was provided, the inferred joint angle and torque profiles were biased toward strong force characteristics, with joint angles distributed around 2.75 rad and torques between 0.10 and 0.20 Nm. This result indicates that the Bi-ACT model is unable to differentiate between the ``softly'' and ``strongly'' actions, thereby failing to encode the nuanced force requirements for varying task conditions. These findings underscore the necessity of incorporating natural language cues to guide force modulation, ultimately enabling a more precise and context-aware imitation learning process.

We tested action-oriented prompts, ``softly grasp the cup'' and ``strongly grasp the cup'', to determine whether Bi-LAT could replicate the training distributions shown in Fig.~\ref{fig:trainingdata}.
While the overall task was executed successfully, time series graph and histogram analyses revealed varying levels of differentiation among the language models.
For example, as shown in Fig.~\ref{fig:cup_exp_all}(b), DistilBERT produced a joint angle distribution for ``softly grasp the cup'' between 2.65 and 2.7 rad, while for ``strongly grasp the cup'' the joint angles exceeded 2.7 rad, and corresponding torques were near 0 Nm and around 0.20 Nm, respectively. 
However, compared to the training data, the motions generated for ``softly grasp the cup'' exhibited higher than expected angles and torques.
In contrast, the results shown in Fig.~\ref{fig:cup_exp_all}(c) for ModernBERT and Fig.~\ref{fig:cup_exp_all}(d) for CLIP demonstrated some differentiation in the maximum values, but the overall distributions did not clearly distinguish between the two conditions.
Most notably, Fig.~\ref{fig:cup_exp_all}(e) shows that SigLIP provided the clearest separation, with the inference data for each instruction closely matching the training distributions.

These findings confirm that incorporating natural language cues into a bilateral control-based imitation learning framework is essential for precise force modulation. By aligning with the distributions in the training data, SigLIP-based Bi-LAT can effectively map language instructions to the appropriate joint angle and torque profiles.

\section{Bimanual Experiment}
\begin{figure}[t] 
    \centering \includegraphics[keepaspectratio, width=1\linewidth]{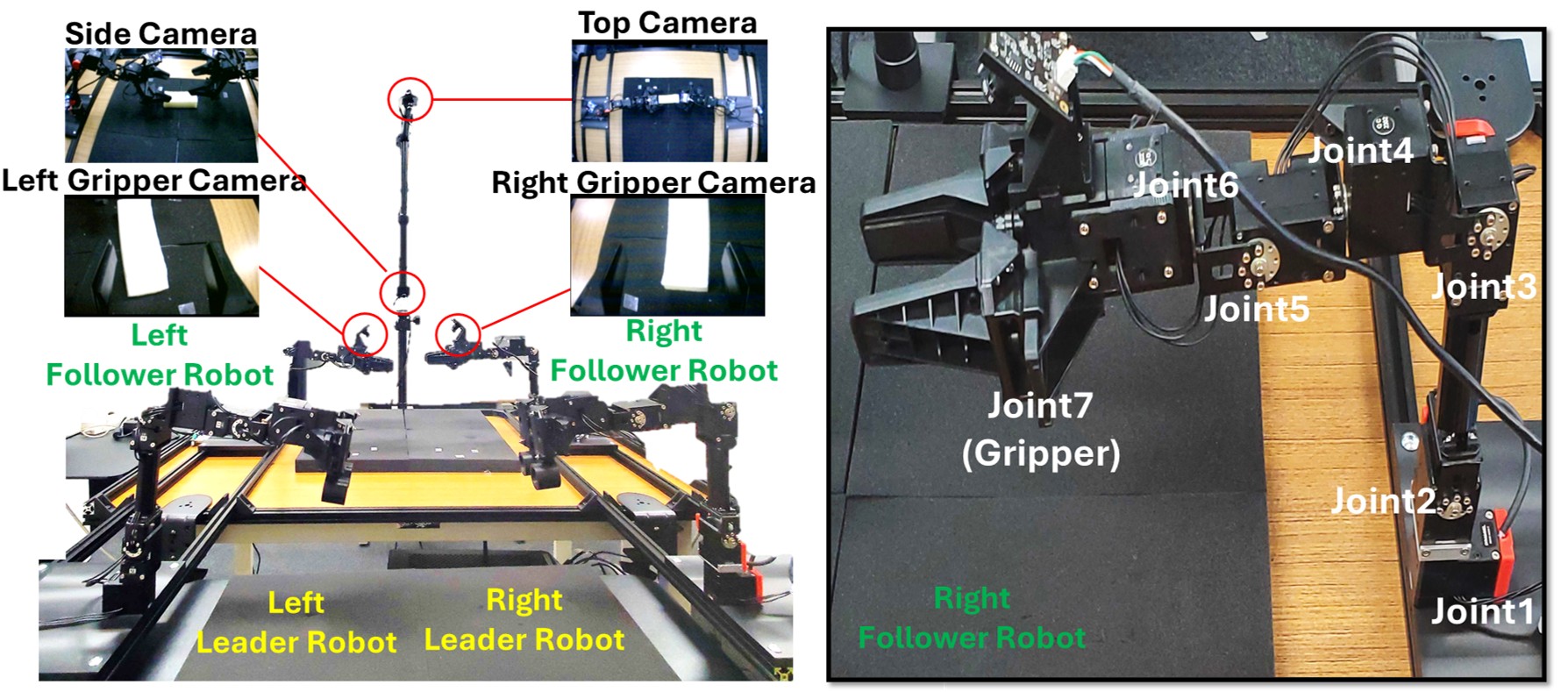} 
    \caption{Bimanual Experimental Environment} 
    \label{fig:ex2_hardware} 
\end{figure}
\begin{figure*}[t] 
    \centering 
    \includegraphics[keepaspectratio, width=1\linewidth]{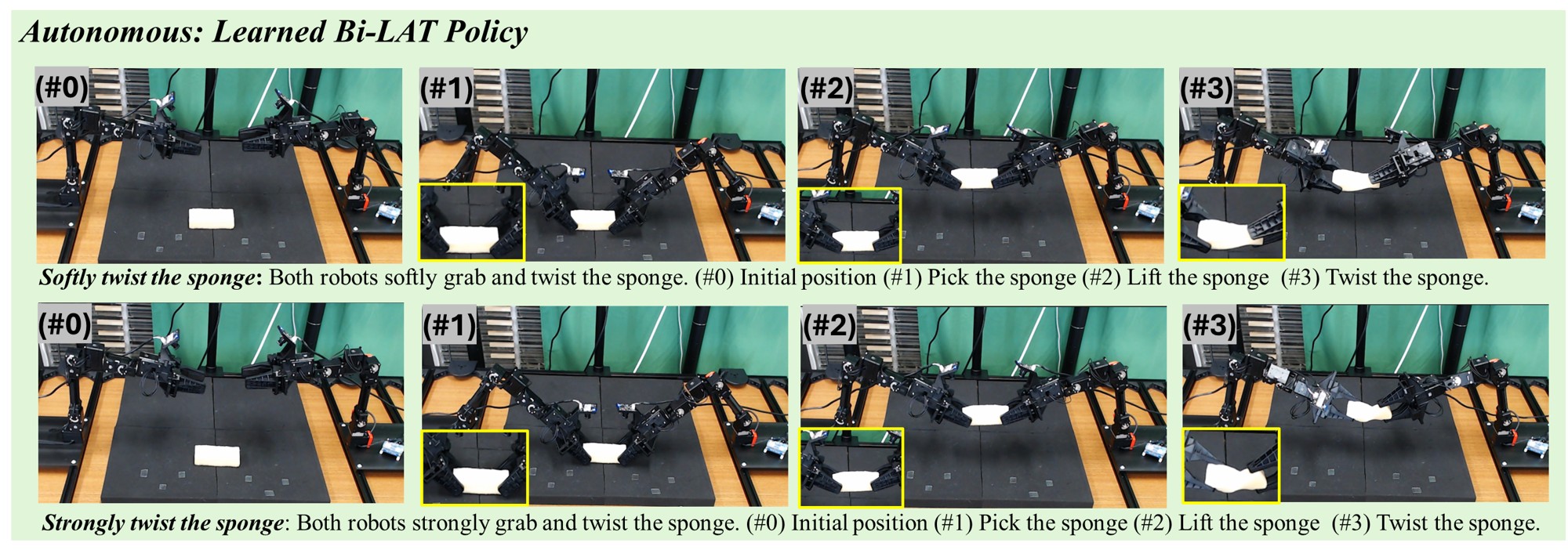} 
    \caption{Results of Sponge-Twisting Task using ALPHA-$\alpha$} 
    \label{fig:ex2_autonomous} 
\end{figure*}
\subsection{Hardware}
As shown in Fig.~\ref{fig:ex2_hardware}, the bimanual robots ALPHA-$\alpha$\cite{kobayashi2025alpha} were used for experiments in bimanual robotic manipulation.
A total of four robots were employed, including two leader robots operated by the human operator and two follower robots. Each robot features 6 degrees of freedom (DOF) for versatile movement, and an additional DOF for the gripper, utilizing a total of 7 motors for its operation.
The bilateral control cycle was set to 1000 Hz for precise movement and data collection of joint angle, velocity, and torque.
Furthermore, four RGB cameras were positioned overhead, on the side, and on the right and left gripper joints of the follower robots to capture visual observations.

\subsection{Experimental Conditions}
To evaluate the effectiveness of Bi-LAT with SigLIP in linking natural language instructions to robotic motor behavior, we ran a sponge-twisting task via bilateral teleoperation.

As illustrated in Fig.~\ref{fig:ex2_datacollection}, the sponge-twisting task consists of three stages:
\begin{itemize}

\item \textit{Grab}: The robots open their grippers, move forward, and close them to grasp the sponge.
\item \textit{Lift}: The sponge is lifted from the table surface.
\item \textit{Twist}: The sponge is twisted while being held.
\end{itemize}

To examine the effect of varying force requirements, we collected demonstrations for two distinct instructions: 
\begin{itemize}
    \item \textit{``Softly twist the sponge''} (lighter grip force)
    \item \textit{``Strongly twist the sponge''} (firmer grip force)
\end{itemize}

Because the sponge was deformable, these instructions induced noticeably different torque profiles, particularly during the grabbing and twisting phases. 

\subsection{Training Setup}
Joint angles, angular velocities, and torques were recorded from both the leader and follower robots using the 4-channel bilateral control system, operating at the control frequency of 1000 Hz.
This resulted in 21-dimensional joint data (3 values × 7 joints) per robot, and 84-dimensional data in total across both robots, capturing detailed motor-level dynamics essential for imitation learning.
Simultaneously, RGB images were captured from four cameras at a frequency of 100 Hz.
\begin{figure}[t] 
    \centering 
    \includegraphics[keepaspectratio, width=1\linewidth]{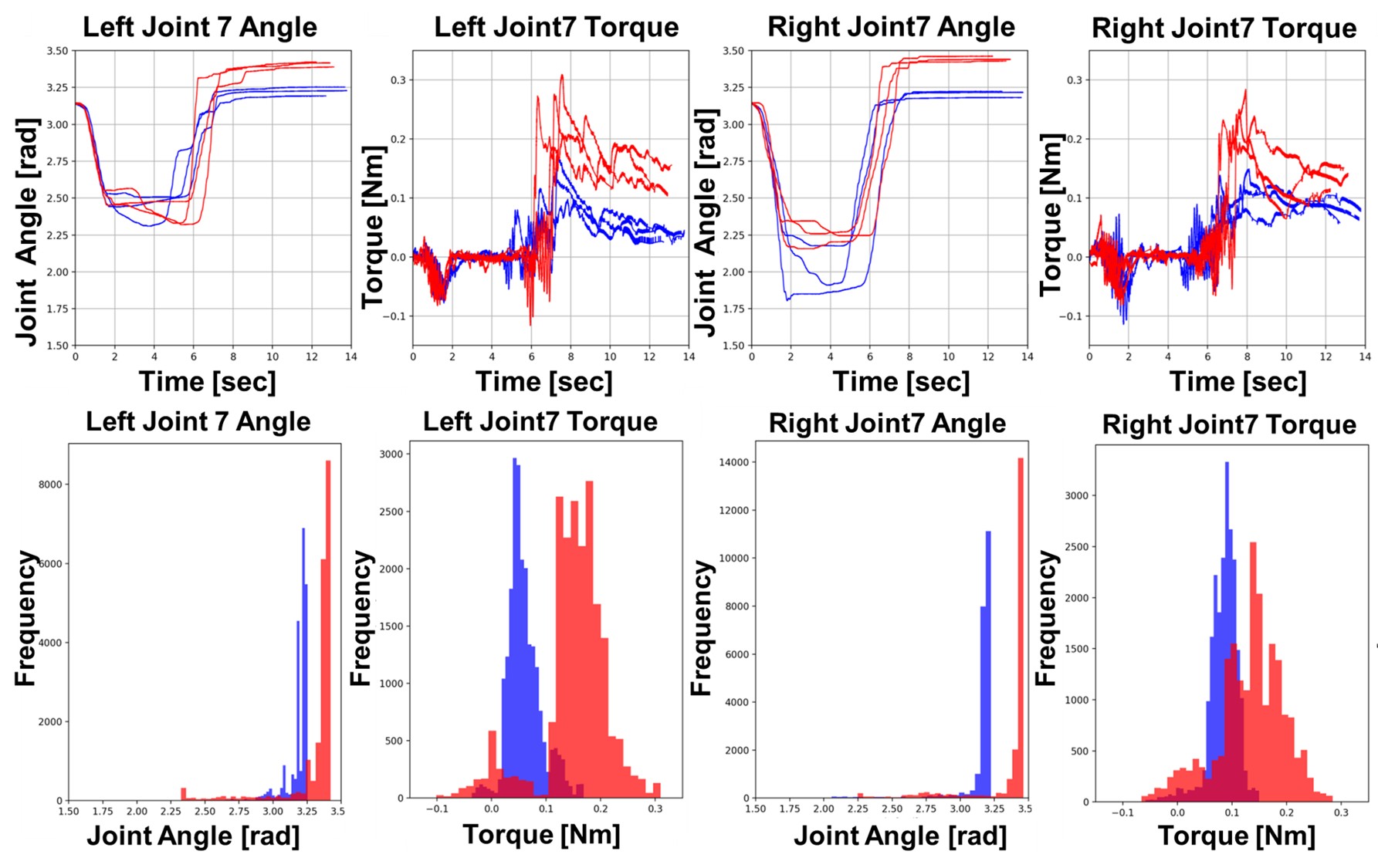} 
    \caption{Training Data of Follower Robots} 
    \label{fig:spg_train_data} 
\end{figure}
\begin{table}[t]
    \centering
    \caption{Sponge-Twisting Success Rates by Stage (Bi-LAT with SigLIP)}
    \label{tab:twist_sponge_result}
    \scalebox{1}{
    \begin{tabular}{ccccc}
        \hline
        \textbf{Instruction} & \textbf{Pick} & \textbf{Lift} & \textbf{Twist} & \textbf{Overall} \\ \hline\hline
        Softly  & 5/5(100\%) & 5/5(100\%) & 5/5(100\%) & 5/5(100\%) \\
        Strongly  & 5/5(100\%) & 5/5(100\%) & 4/5(80\%) & 4/5(80\%) \\ \hline
    \end{tabular}
    }
\end{table}

As shown in Fig.~\ref{fig:spg_train_data}, each demonstration was paired with a natural language instruction; among the six demonstrations, three involved ``strongly twist the sponge'', while the other three involved ``softly twist the sponge.''
The histograms were generated from data captured after six seconds, corresponding to the interval in which the sponge was actively grasped.
Under the ``softly twist the sponge'' instruction, most joint angles were distributed between 3.0 and 3.25 rad, with torques between 0 and 0.10 Nm. 
Under the ``strongly twist the sponge'' instruction, joint angles concentrated between 3.25 and 3.5 rad, with torques between 0.10 and 0.30 Nm.

By applying DABI, we expanded the original six demonstrations into 60 by downsampling the 1000 Hz control data to 100 Hz, thereby increasing the dataset size by a factor of ten.
Based on this dataset, we trained Bi-LAT using the SigLIP text encoder and configured the model with 4 encoder layers and 7 decoder layers.
\begin{figure}[t] 
    \centering 
    \includegraphics[keepaspectratio, width=1\linewidth]{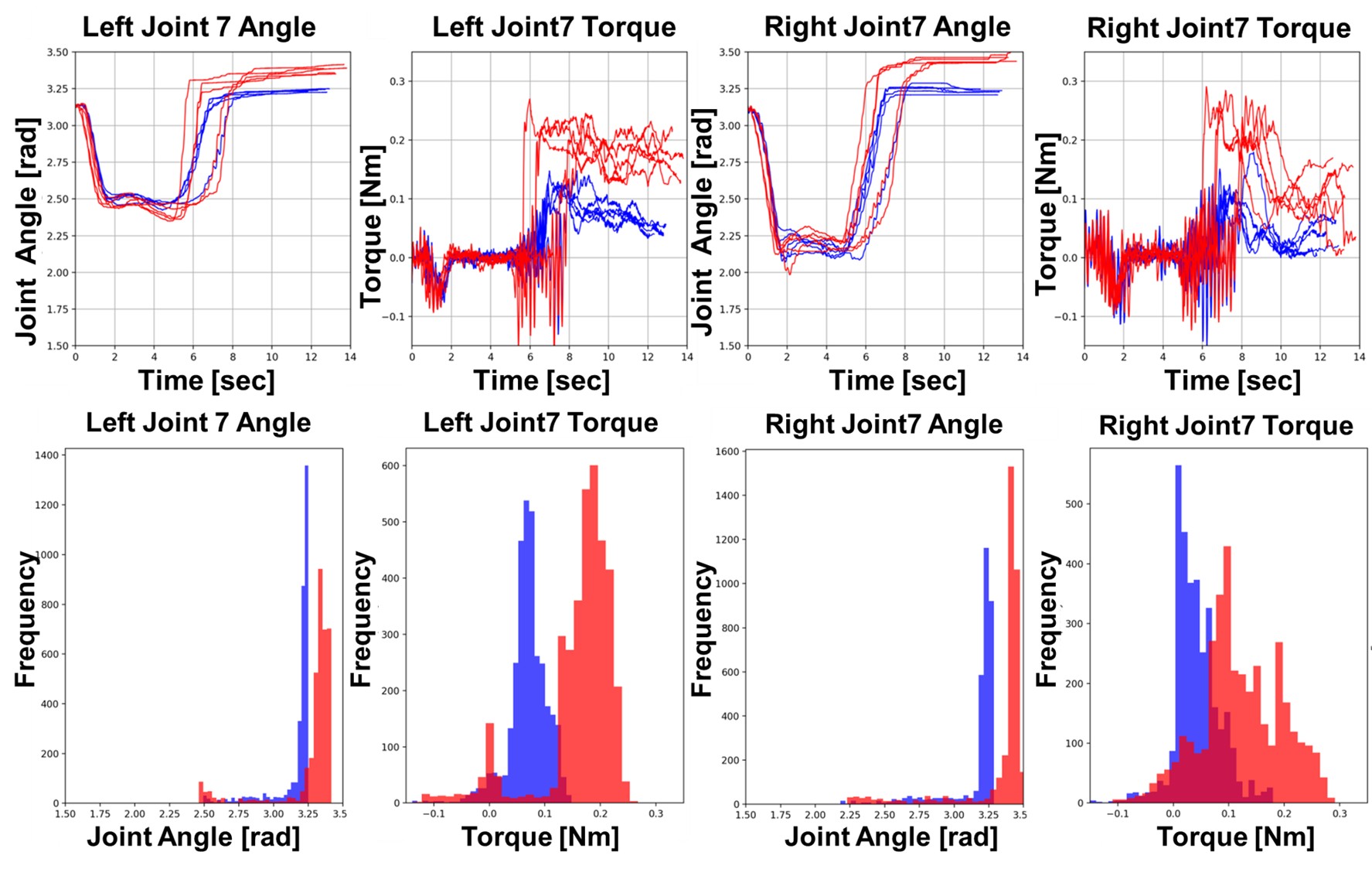} 
    \caption{Results of Follower Robots via Bi-LAT} 
    \label{fig:spg_ex_data} 
\end{figure}
\subsection{Experiment Result}

As shown in Fig.~\ref{fig:ex2_autonomous},  we conducted autonomous executions of the sponge-twisting task under two distinct instructions: ``softly twist the sponge'' and ``strongly twist the sponge.'' As summarized in Table~\ref{tab:twist_sponge_result}, the system achieved a 100\% success rate across all Pick, Lift, Twist phases under the soft twist instruction. In contrast, under the strong twist instruction, Pick and Lift both succeeded in 100\% of trials, whereas Twist succeeded in four out of five trials (80\%). A closer examination of the failure case revealed that the sponge’s higher internal resistance caused the robot to lose a secure grip during twisting.

During autonomous execution, as shown in Fig.~\ref{fig:spg_ex_data}, the Bi-LAT model followed comparable spatial trajectories for both ``soft'' and ``strong'' conditions, yet produced markedly different torque profiles. 
Red lines represent strongly grasped-and-twisted runs, while blue lines indicate softly grasped-and-twisted runs. Although the overall motions share similar paths, torque values diverge sharply during grasping and twisting. Furthermore, histograms of both joint angles and torques confirm that the two conditions form distinct distributions that closely match their respective training data. These results demonstrate the model's ability to modulate force based on natural language instructions rather than reproducing a uniform trajectory, showcasing the value of integrating language cues into a bilateral control-based imitation learning framework for nuanced force control.

\section{CONCLUSIONS}\label{sec:AW}
This paper presented Bi-LAT, a novel framework that integrates natural language processing into bilateral control-based imitation learning to achieve intuitive force modulation in robotic manipulation.
Building on the foundation of Bi-ACT, Bi-LAT leverages visual, proprioceptive, and linguistic data to generate actions that accurately reflect the nuanced force levels specified by human operators. 

Experiments in both unimanual and bimanual settings confirmed that Bi-LAT reliably reproduces a range of force intensities, particularly when using SigLIP as the text encoder. This precision in force modulation, integrated with language-based commands and haptic feedback, demonstrates the framework's robustness in adapting to dynamic manipulation tasks. Our results underscore the potential of combining bilateral control-based imitation learning with natural language cues to enhance the adaptability and intuitiveness of robotic manipulation. The hybrid approach enhances complex object manipulation and promotes seamless human-robot collaboration across service, industrial, and assistive robotics.

Future work will address more diverse object properties including rigid and deformable materials, incorporate varying task difficulties, refine real-time language interactions, and explore large-scale datasets for enhanced generalization.


\bibliographystyle{IEEEtran}

\end{document}